\documentclass[]{include/trbunofficial}
\usepackage{amsmath,amsfonts}
\usepackage{array}
\usepackage{textcomp}
\usepackage{stfloats}
\usepackage{url}
\usepackage{verbatim}
\usepackage{graphicx}
\usepackage{array}
\newcolumntype{P}[1]{>{\centering\arraybackslash}p{#1}}
\usepackage{fullpage}
\usepackage{graphicx}
\usepackage{stmaryrd}
\usepackage{amsthm}
\usepackage{amssymb}
\usepackage{courier}
\usepackage{pifont}
\usepackage{color}
\usepackage{verbatim}
\usepackage{appendix}
\usepackage[hidelinks]{hyperref}
\usepackage[scr=boondoxo]{mathalfa}
\usepackage[section]{placeins}
\graphicspath{ {figures/} }
\usepackage{float}
\usepackage[labelfont=bf, textfont=bf]{caption}
\usepackage{balance}
\usepackage{breqn}
\usepackage{accents}
\usepackage{algorithm}
\usepackage{algorithmic}
\usepackage{booktabs}
\usepackage{threeparttable}
\usepackage{makecell}
\usepackage{multirow}
\usepackage{pdflscape}

\usepackage[font=small]{subcaption}

\titleformat{\section}{\bfseries}{\thesection.}{1em}{\uppercase}
\titleformat{\subsection}{\bfseries}{\thesubsection.}{1em}{}
\titleformat{\subsubsection}{\itshape}{\thesubsubsection.}{1em}{}

\title{A Comparative Study of Spline-Based Trajectory Reconstruction Methods Across Varying Automatic Vehicle Location Data Densities}

\author{%
  \textbf{Jake Robbennolt}\\
    University of Texas at Austin\\
    Department of Civil, Architectural and Environmental Engineering\\
    301 E. Dean Keeton St. Stop C1761, 
    Austin TX 78712, USA\\
    Email: \texttt{jr73453@utexas.edu}\\
  \hfill\break%
  \textbf{Sirajum Munira}\\
    Center for Transportation Research,\\ The University of Texas at Austin\\
    3925 W Braker Ln
    Austin TX 78712, USA\\
	Email: \texttt{munira\_silvy@utexas.edu}\\
  \hfill\break%
  \textbf{Stephen D. Boyles}\\
    University of Texas at Austin\\
    Department of Civil, Architectural and Environmental Engineering\\
    301 E. Dean Keeton St. Stop C1761, 
    Austin TX 78712, USA\\
	Email: \texttt{sboyles@austin.utexas.edu}
}

\TotalWords{6491}

\begin{document}
\maketitle

\section*{Abstract}
Automatic vehicle location (AVL) data offers unprecedented insights into transit dynamics, but its effectiveness is often hampered by inconsistent update frequencies, necessitating trajectory reconstruction. This research evaluates 13 trajectory reconstruction methods, including several novel approaches, using high-resolution AVL data from Austin, Texas. We examine the interplay of four critical factors -- velocity, position, smoothing, and data density -- on reconstruction performance. A key contribution of this study is the evaluation of these methods across both sparse and dense datasets, providing insights into a critical trade-off between accuracy and resource allocation. Our evaluation framework combines traditional mathematical error metrics for positional and velocity with practical considerations, such as physical realism (e.g., aligning velocity and acceleration with stopped states, deceleration rates, and speed variability). In addition, we provide insight into the relative value of each method in calculating realistic metrics for infrastructure evaluations. Our findings indicate that velocity-aware methods consistently outperform position-only approaches. Interestingly, we discovered that smoothing-based methods can degrade overall performance in complex, congested urban environments, although enforcing monotonicity remains critical. The velocity constrained Hermite interpolation with monotonicity enforcement (VCHIP-ME) yields optimal results, offering a balance between high accuracy and computational efficiency. Its minimal overhead makes it suitable for both historical analysis and real-time applications, providing significant predictive power when combined with dense datasets. These findings offer practical guidance for researchers and practitioners implementing trajectory reconstruction systems and emphasize the importance of investing in higher-frequency AVL data collection for improved operational analysis.

\hfill\break%
\noindent\textit{Keywords}: Automatic Vehicle Location Data, Trajectory Reconstruction 
\newpage

\section{Introduction}
The availability of high temporal and geographic resolution data from bus automatic vehicle location (AVL) and automated passenger count (APC) systems has opened new avenues for researchers and practitioners to gain insights into transit operations. While APC data is commonly used to analyze average vehicle speeds, headways, and boarding/alighting patterns across routes, AVL data provides a more detailed understanding of transit dynamics. AVL data has many established use cases including evaluating where slowdowns occur between stops, updating signal timings to improve operations, and pinpointing potential safety concerns \citep{huang_reconstructing_2023}. However, inconsistent update frequencies \citep{cui_autonomous_2003, sirisombat_electronic_2024} can create challenges in forming a continuous trajectory, particularly in congested downtown areas with closely spaced traffic signals. For many applications, understanding what happens between two consecutive AVL data points is crucial, highlighting the need for trajectory reconstruction. 

Trajectory reconstruction is a well-studied topic in various domains \citep{feng_novel_2021,guo_improved_2021,wang_general_2022}; however, applying it to transit vehicles presents unique challenges and opportunities due to their specific operational characteristics. Generic methods struggle to adapt to rapid fluctuations in AVL data caused by frequent stops and interactions with traffic signals. In addition, transit vehicle operate with large headways, meaning each trajectory is essentially independent (unlike applications such as traffic state estimation \citep{ni_trajectory_2008,sun_vehicle_2013,robbennolt_data-driven_2024} or origin-destination pattern estimation \citep{rao_origin-destination_2018, mo_estimating_2020}). In addition, the more aggregate origin-destination flows and travel times can be more easily found from APC data \citep{patnaik_estimation_2004, kuipers_passenger_2022, bin_zulqarnain_addressing_2023}. On the positive side, transit vehicles operate on fixed routes simplifying the process of matching data to geographic coordinates. Additionally, AVL data often includes velocity information, which can significantly enhance the quality of trajectory reconstruction \citep{yu_curve-based_2004}. 

As connected vehicle technology progresses and velocity data become increasingly available, trajectory reconstruction methods must be adapted. Traditional spline-based methods such as piecewise cubic Hermite interpolating polynomials (PCHIP) have been widely used for trajectory interpolation due to their shape-preserving properties and computational efficiency \citep{fritsch_monotone_1980, fritsch_method_1984, sun_vehicle_2016, li_cubic_2022}. More sophisticated approaches including velocity-constrained splines, smoothing methods, and state estimation techniques such as Kalman and particle filters have been developed to address measurement noise and ensure physically realistic vehicle dynamics \citep{yu_curve-based_2004, ni_trajectory_2008, cao_v-spline_2021, wei_particle_2021, dong_enhanced_2024}, though these filtering approaches often require dynamic model assumptions and parameter tuning that may not generalize well across different transit operations. In addition, some recent deep learning approaches have been proposed, though these typically require very large datasets for training, significant computational resources, and are less explainable and generalizable across systems and routes \citep{chen_modeling_2021,zhao_high-fidelity_2022,wang_general_2022}. Moreover, many existing methods fall short in complex urban environments where buses experience frequent accelerations and decelerations due to traffic signals, passenger stops, and congestion. In such environments, simple interpolation methods (like linear interpolation) may fail to capture the underlying vehicle dynamics, leading to unrealistic velocity profiles and potentially erroneous conclusions in subsequent analyses. Compounding this, the trade-offs between data collection frequency, computational efficiency, and reconstruction accuracy remain poorly understood, limiting the practical guidance available to transit agencies implementing these systems.

Despite the extensive literature on trajectory interpolation, several gaps remain. First, most existing studies focus on either position-only or velocity-only approaches. In particular, \citet{cao_v-spline_2021} developed a promising velocity-aware smoothing spline algorithm, but compared it only with position-only interpolation approaches. More recently, \citet{huang_reconstructing_2023} studied the same problem of interpolating vehicle trajectories and argued that the resultant interpolation should create a monotonically increasing trajectory that is continuous and at least once differentiable. They suggest that the trajectory should be made up of composite cubic polynomials since that is the lowest order curve that can approximate the trajectory \citep{huang_reconstructing_2023}. However, their dataset only included location data, so they did not study any methods that used velocities.

Building on current research, this study addresses several gaps that limit the practical deployment of trajectory reconstruction in transit systems. First, we examine the interplay between position, velocity, and smoothing on reconstruction performance to provide practitioners a guide to choosing an approach that meets their needs and data requirements. Second, we quantify how data density impacts reconstruction quality for transit, a previously underexplored area. Since data collection and storage can be expensive, this analysis can help practitioners allocate effort to appropriate levels of data collection based on their specific needs. Third, our evaluation goes beyond simple mathematical error metrics to consider practical implications for infrastructure analysis and operational decision-making by comparing metrics such as deceleration rates and speed variability and by validating trajectories against realistic profiles. Since trajectories are typically an intermediate processing step rather than the end goal, this research bridges the gap by connecting reconstruction quality to specific performance metrics that are used in practice for infrastructure evaluations. Fourth, we characterize the computational trade-offs that determine whether methods can operate in real-time systems, addressing a critical barrier to practical implementation. Finally, we make technical contributions by introducing several novel velocity-aware methods, with a particular focus on monotonicity enforcement.

This paper is organized as follows: Section \ref{sec:methods} provides background and mathematical details on each of the thirteen smoothing methods studied. At the end of that section we also present a theoretical comparison of each of the methods in terms of construction approach, monotonicity enforcement, differentiability, data requirements, and the number of parameters that must be tuned. Section \ref{sec:case_study} presents background on the case study including details about the AVL data taken from rapid transit routes through downtown Austin, TX. Section \ref{sec:results} presents the results of the interpolation evaluation in three subsections corresponding to error-based metrics, physics-based metrics, and metrics related to practical implementation. Finally, Section \ref{sec:conclusion} concludes the paper with final recommendations for the best trajectory smoothing approaches and recommendations for further study. 

\section{Trajectory Reconstruction Methods}
\label{sec:methods}
We extend the work of \citet{huang_reconstructing_2023} by considering both position and velocity in trajectory reconstruction. We implemented four approaches using position alone, and nine using both position and velocity. Each method is described in this section. 

\subsection{Position-Based Interpolation Methods}
Each of these methods uses only position-based information. This is beneficial when velocity data is unreliable or expensive to collect. 

\subsubsection{Linear Segment Interpolation (LSEG)}
The simplest method for trajectory reconstruction connects adjacent data points with straight lines. For a set of time-distance observations $(t_i, x_i)$, linear interpolation produces a continuous but non-differentiable trajectory function:
\begin{align}
\hat x(\hat t) &= x_i + \frac{x_{i+1} - x_i}{t_{i+1} - t_i}(\hat t - t_i), \quad t_i \leq \hat t \leq t_{i+1}.
\end{align}

Velocity is approximated with a piecewise constant velocity profile:
\begin{align}
\hat v(\hat t) &= \frac{x_{i+1} - x_i}{t_{i+1} - t_i}, \quad t_i \leq \hat t < t_{i+1}.
\end{align}

LSEG guarantees monotonicity if the initial vehicle locations have been preprocessed to ensure they are already monotonic. It is also computationally efficient, but produces unrealistic instantaneous velocity changes at observation points and can be sensitive to outliers in the data.

\subsubsection{Piecewise Cubic Hermite Interpolation (PCHIP)}
PCHIP creates a smooth trajectory by fitting cubic polynomials between adjacent data points while preserving monotonicity and ensuring first-derivative continuity.

For each interval $[t_i, t_{i+1}]$, the position function is defined as:
\begin{align}
\hat x(\hat t) &= h_{00}(s)x_i + h_{10}(s)(t_{i+1}-t_i)m_i + h_{01}(s)x_{i+1} + h_{11}(s)(t_{i+1}-t_i)m_{i+1},
\label{eq:x_pchip}
\end{align}
where:
\begin{align}
    s &= \frac{\hat t-t_i}{t_{i+1}-t_i},\\
    h_{00}(s) &= 2s^3-3s^2+1,\\
    h_{10}(s) &= s^3-2s^2+s,\\
    h_{01}(s) &= -2s^3+3s^2,\\
    h_{11}(s) &= s^3-s^2,
\end{align}
and $m_i$ are the tangents for each interval (which should be consistent at shared endpoints to ensure continuity). Using the Fritsch–Carlson method to choose these tangents \citep{fritsch_monotone_1980, fritsch_method_1984}, we ensure monotonicity. The velocity can be obtained directly by differentiating the position function:
\begin{align}
\hat v(\hat t) &= \frac{d}{dt}\left[\hat x(\hat t)\right]. \label{eq:v_pchip}
\end{align}

Algorithm \ref{alg:fritsch-carlson} shows the Fritsch-Carlson approach to ensuring monotonic tangents. Essentially, this approach initializes the tangents to be $m_i = \delta_i$ where: 
\begin{align}
    \delta_i = \frac{x_{i+1} - x_i}{t_{i+1} - t_i}. \label{eq:fc1}
\end{align}
Then, \citet{fritsch_monotone_1980} determined that monotonicity can be guaranteed if $m_i$ satisfies:
\begin{align}
    \alpha_i = \frac{m_i}{\delta_i}, \\
    \beta_i = \frac{m_{i+1}}{ \delta_i}, \\
    \alpha_k^2 + \beta_k^2 \leq 9.
\end{align}

If this constraint is violated, then, set 
\begin{align}
    \tau_i = 3 / \sqrt{\alpha_i^2 + \beta_i^2}, \\
    m_i = \tau_i \alpha_i \delta_i, \\
    m_{i+1} = \tau_i \beta_i \delta_i.
    \label{eq:fcn}
\end{align}
When all $m_i$ are created in this way, they will be equal for all intervals, allowing equations \eqref{eq:x_pchip} and \eqref{eq:v_pchip} to be applied directly. 

\begin{algorithm}
\caption{Fritsch-Carlson Monotonic Hermite Interpolation}
\label{alg:fritsch-carlson}
\begin{algorithmic}
    \REQUIRE $[T,X,V] = [(t_1, x_1, v_1), (t_2, x_2, v_2), \ldots, (t_n, x_n, v_n)]$ where $t_1 < t_2 < \cdots < t_n$
    \FOR{$i = 1, 2, \ldots, n-1$}
        \STATE $\delta_i \leftarrow \frac{x_{i+1} - x_i}{t_{i+1} - t_i}$ \COMMENT{Secant slopes}
    \ENDFOR
    \STATE $m_1 \leftarrow \delta_1$, $\qquad m_n \leftarrow \delta_{n-1}$
    \STATE $m_i \leftarrow \frac{\delta_{i-1} + \delta_i}{2}$ for $i =  2, \ldots, n-1$ 
    \FOR{$k = 1, 2, \ldots, n-1$}
        \IF{$|\delta_k| < \epsilon$} 
            \STATE $m_k \leftarrow 0$, $m_{k+1} \leftarrow 0$ \COMMENT{Nearly flat interval}
        \ELSE
            \STATE $\alpha_k \leftarrow m_k / \delta_k$, $\beta_k \leftarrow m_{k+1} / \delta_k$
            
            \IF{$\alpha_k^2 + \beta_k^2 > 9$} 
                \STATE $\tau_k \leftarrow 3 / \sqrt{\alpha_k^2 + \beta_k^2}$, 
                $\qquad m_k \leftarrow \tau_k \cdot \alpha_k \cdot \delta_k$, 
                $\qquad m_{k+1} \leftarrow \tau_k \cdot \beta_k \cdot \delta_k$ \COMMENT{Circle constraint}
            \ENDIF
        \ENDIF
    \ENDFOR
    \STATE $\hat f \leftarrow [   x(\hat t) = h_{00} \cdot x_k + h_{10} \cdot h \cdot m_k + h_{01} \cdot x_{k+1} + h_{11} \cdot h \cdot m_{k+1}, v(\hat t) = \frac{d}{dt}\hat x(\hat t) ]$ 
    
    \RETURN $\hat f$ \COMMENT{Return interpolation function for prediction at new time points $\hat t$}
\end{algorithmic}
\end{algorithm}

\subsubsection{Local Regression Smoothing (LOCREG)}
To account for measurement errors in the observed data, we implement local regression smoothing \citep{cleveland_robust_1979, wand_kernel_1994}. LOCREG fits a polynomial function to each point using weighted least squares regression, with weights determined by a kernel function that emphasizes nearby points:
\begin{align}
\min \sum_{i=1}^{n} w_i(\hat t) \left(x_i - f(t_i)\right)^2,
\end{align}
where $w_i(\hat t)$ is the weight assigned to observation $i$ when estimating at evaluation time $\hat t$, and $f(\hat t)$ is a cubic polynomial. At each evaluation point, the algorithm fits a cubic polynomial to nearby data points within a neighborhood defined by the $k$ nearest neighbors. The weights are determined by the tricube kernel function:
\begin{align}
w(u) = \begin{cases}
(1 - |u|^3)^3 & \text{if } |u| \leq 1, \\
0 & \text{if } |u| > 1,
\end{cases}
\end{align}
where $u = (t_i - \hat t)/h$ represents the normalized distance from the evaluation point $\hat t$ to each data point $t_i$ and $h$ is the bandwidth parameter corresponding to the distance to the k-th nearest neighbor \citep{cleveland_robust_1979}.

\subsubsection{Combined Local Regression Smoothing with Piecewise Cubic Hermite Interpolation (LOCREG-PCHIP)}
Following \citet{huang_reconstructing_2023}, we implement a combined approach that leverages the strengths of both LOCREG and PCHIP. In this two-stage process, we first apply the local polynomial regression (i.e. for every point $(t_i, x_i)$ we determine a smoothed intermediate estimate $(t_i,y_i)$). The second stage enforces strict monotonicity through a sequential correction procedure, ensuring that each position $y_i$ satisfies $y_i \geq y_{i-1}$. Finally, we apply monotonic PCHIP interpolation to map from these monotonicity-corrected points to the target interpolation locations $(\hat t, \hat x)$ (see Algorithm \ref{alg:locreg-pchip}).  

\begin{algorithm}
\caption{LOCREG-PCHIP Interpolation}
\label{alg:locreg-pchip}
\begin{algorithmic}
    \REQUIRE $[T,X] = [(t_1, x_1), (t_2,x_2), \ldots, (t_n,x_n)]$ where $t_1 < t_2 < \cdots < t_n$ and $x_1 \leq x_2 \leq \cdots \leq x_n$
    \REQUIRE $k$ \COMMENT{Neighborhood size}
    \STATE $f \leftarrow \text{LOCREG}(T,X, k)$ \COMMENT{Fit LOCREG model with neighborhood $k$}
    \FOR{$i = 1, 2, \ldots, n$}
        \STATE $y_i \leftarrow f_{\text{LOCREG}}(t_i)$ \COMMENT{Predict modified locations $y_i$ at original time points $t_i$}
        \IF{$i > 1$ \AND $y_i < y_{i-1}$}
            \STATE $y_i \leftarrow y_{i-1}$ \COMMENT{Enforce monotonicity}
        \ENDIF
    \ENDFOR
    \STATE $\hat f \leftarrow \text{PCHIP}(T, Y)$ \COMMENT{Fit monotonic PCHIP spline}
    \RETURN $\hat f$ \COMMENT{Return interpolation function for prediction at new time points $\hat t$}
\end{algorithmic}
\end{algorithm}

This hybrid approach produces trajectories that are smooth, monotonic, and differentiable while accounting for measurement errors. 

\subsection{Velocity-Aware Interpolation Methods}
While Huang et al. (2023) demonstrated that the LOCREG-PCHIP approach can be especially beneficial, they assumed that no velocity data was available. Where that data is available, it can be very helpful to utilize it to create trajectories which better match the curvature of realistic drivers. 

\subsubsection{Linear Velocity Matching Interpolation (LVMI)}
For each interval $[t_i, t_{i+1}]$, we construct two linear functions that match the observed velocities at each endpoint:
\begin{align}
x^{(i)}(\hat t) &= x_i + v_i(\hat t - t_i), \\
x^{(i+1)}(\hat t) &= x_{i+1} + v_{i+1}(\hat t - t_{i+1}).
\end{align}

The intersection time $t_{int}$ is determined by setting $x^{(i)}(t_{int}) = x^{(i+1)}(t_{int})$ and solving:
\begin{align}
t_{int} = \frac{x_{i+1} - x_i + v_{i+1}t_{i+1} - v_it_i}{v_{i+1} - v_i}.
\end{align}

If the intersection of these lines falls within the interval, we use $\hat x(\hat t) = x^{(i)}(\hat t)$ for $\hat t \leq t_{int}$ and $\hat x(\hat t) = x^{(i+1)}(\hat t)$ for $\hat t > t_{int}$. Otherwise, we use the function from the nearest endpoint.

This approach guarantees that the reconstructed trajectory matches both the observed positions and velocities at data points, but it produces discontinuous velocity profiles.

\subsubsection{Velocity Constrained Hermite Interpolation (VCHIP)}
For each interval $[t_i, t_{i+1}]$, we define a cubic polynomial:
\begin{align}
\hat x(\hat t) &= a_0 + a_1(\hat t-t_i) + a_2(\hat t-t_i)^2 + a_3(\hat t-t_i)^3.
\label{eq:x_vchip}
\end{align}

The coefficients are determined by four constraints: matching the position and velocity at both endpoints. This yields:
\begin{align}
a_0 &= x_i, \\
a_1 &= v_i, \\
a_2 &= \frac{3(x_{i+1} - x_i)}{h^2} - \frac{2v_i + v_{i+1}}{h}, \\
a_3 &= \frac{-2(x_{i+1} - x_i)}{h^3} + \frac{v_i + v_{i+1}}{h^2},
\end{align}
where $h = t_{i+1} - t_i$ is the time interval duration.

The velocity function is obtained by differentiating:
\begin{align}
\hat v(\hat t) &= a_1 + 2a_2(\hat t-t_i) + 3a_3(\hat t-t_i)^2.
\label{eq:v_vchip}
\end{align}

Equations \eqref{eq:x_vchip} and \eqref{eq:v_vchip} are equivalent to equations \eqref{eq:x_pchip} and \eqref{eq:v_pchip}, where the tangents $m_i$ have been replaced by the velocities $v_i$. However, there is no guarantee that the VCHIP is monotonically increasing, a major drawback. 

\subsubsection{Velocity Constrained Hermite Interpolation with Monotonicity Enforcement (VCHIP-ME)}
While VCHIP ensures smooth trajectories that match observed position and velocity data, it does not guarantee monotonicity. VCHIP-ME addresses this limitation by applying PCHIP-style monotonicity constraints.

The method operates in two stages: first, constraining the observed velocities to ensure monotonicity (equations \eqref{eq:fc1} through \eqref{eq:fcn}). Then, this method follows exactly Algorithm \ref{alg:fritsch-carlson} (standard cubic Hermite interpolation), but instead of initializing the tangent slopes to be the secant slopes, they are initialized to be the velocities ($m_i = v_i$). Though somewhat more constrained than the VCHIP method, this approach combines the smoothness advantages of cubic Hermite interpolation with the monotonicity guarantees essential for realistic vehicle trajectory reconstruction. 

\subsubsection{Blended Piecewise Cubic Hermite Interpolation and Velocity Constrained Hermite Interpolation (PCHIP-VCHIP)}
VCHIP-ME operates in four stages: computing PCHIP derivatives, blending these with observed velocities, applying final monotonicity constraints to the combined derivatives, and then computing a final monotone Hermite interpolation.
\begin{enumerate}
    \item Calculate PCHIP derivatives $u_i$ for each point.
    \item Combine the PCHIP derivatives with observed velocities using a weighted average: $\tilde v_i = \alpha \cdot v_i + (1-\alpha) \cdot u_i$.
    \item Apply the same Fritsch-Carlson constraints as in VCHIP-ME.
    \item Use standard cubic Hermite interpolation with the constrained blended velocities to estimate final locations and velocities $(\hat t, \hat x, \hat v)$.
\end{enumerate}
The parameter $\alpha \in [0,1]$ is a velocity weight parameter that controls the balance between observed velocities ($\alpha = 1$) and pure PCHIP derivatives ($\alpha = 0$). The velocity weight parameter $\alpha$ allows practitioners to adjust the method based on their confidence in velocity measurements versus position-derived derivatives. Note that steps 3 and 4 are essentially creating a VCHIP-ME interpolation with modified velocities $\tilde v$ (see Algorithm \ref{alg:pchip-vchip}). 

\begin{algorithm}
\caption{PCHIP-VCHIP Interpolation}
\label{alg:pchip-vchip}
\begin{algorithmic}
    \REQUIRE $[T, X, V] = [(t_1, x_1, v_1), (t_2, x_2, v_2), \ldots, (t_n, x_n, v_n)]$ where $t_1 < t_2 < \cdots < t_n$
    \REQUIRE $\alpha \in [0,1]$ \COMMENT{Velocity weight parameter}
    \STATE $g \leftarrow \text{PCHIP\_DERIVATIVES}(T, X)$
    \FOR{$i = 1, 2, \ldots, n$}
        \STATE $u_i = g(t_i)$ \COMMENT{Predict PCHIP derivatives $u_i$ at time points $t_i$}
        \STATE $\tilde{v}_i \leftarrow \alpha \cdot v_i + (1-\alpha) \cdot u_i$ \COMMENT{Create blended velocities $\tilde{V}$}
    \ENDFOR
    \STATE $\hat f \leftarrow \text{VCHIP-ME}(T, X, \tilde{V})$
    \RETURN $\hat f$ \COMMENT{Return interpolation function for prediction at new time points $\hat t$}
\end{algorithmic}
\end{algorithm}

This approach uses a weighted combination of the implied velocities from the location data and the measured velocities, which ensures monotonicity of the final interpolation.

\subsubsection{Local Regression Smoothing with Velocity Constraints (LOCREG-V)}
Building upon the LOCREG approach, LOCREG-V extends local regression smoothing to simultaneously consider both position and velocity observations. This bivariate approach fits separate local polynomial regressions for position and velocity data:
\begin{align}
\min \sum_{i=1}^{n} w_i(\hat t) \left(x_i - f_x(t_i)\right)^2, \\
\min \sum_{i=1}^{n} w_i(\hat t) \left(v_i - f_v(t_i)\right)^2,
\end{align}
where $f_x(\hat t)$ and $f_v(\hat t)$ are cubic polynomials for position and velocity respectively, and both regressions use weight functions $w_i(\hat t)$. 

As with the LOCREG approach, initial smoothed estimates $\tilde{x}_i$ and $\tilde{v}_i$ are found at the original observation times $t_i$ and then used to construct spline functions for both position and velocity that can be evaluated at any target time $\hat t$. Similarly, this approach does not inherently guarantee monotonicity, nor does it even maintain physical consistency between position and velocity. 

\subsubsection{Combined Local Regression Smoothing with Piecewise Cubic Hermite Interpolation with Velocity Constraints (LOCREG-PCHIP-V)}
LOCREG-PCHIP-V combines the noise reduction benefits of bivariate local regression with the monotonicity guarantees and velocity constraints of cubic Hermite interpolation. The algorithm proceeds as follows:

First, we apply local regression simultaneously to both position and velocity data at the original observation points $(t_i, x_i, v_i)$, using the same formulation as LOCREG-V:
\begin{align}
\tilde{x}_i &= f_x(t_i), \\
\tilde{v}_i &= f_v(t_i),
\end{align}
where $f_x$ and $f_v$ are the fitted local regression functions.

Second, we enforce strict monotonicity on the smoothed position estimates:
\begin{align}
y_i = \begin{cases}
\tilde{x}_i & \text{if } i = 1 \text{ or } \tilde{x}_i \geq y_{i-1} \\
y_{i-1} & \text{if } \tilde{x}_i < y_{i-1}
\end{cases}.
\label{eq:locrec_mon1}
\end{align}

Additionally, we adjust the velocity estimates to maintain consistency with the monotonicity-corrected positions:
\begin{align}
u_i = \begin{cases}
\tilde{v}_i & \text{if no monotonicity correction applied} \\
\max\{\frac{y_{i+1} - y_{i-1}}{t_{i+1} - t_{i-1}},0\} & \text{if monotonicity correction applied and } i \in [2, n-1]
\end{cases}.
\label{eq:locrec_mon2}
\end{align}

Finally, we apply VCHIP-ME using the corrected position and velocity estimates $(t_i, y_i, u_i)$ to obtain smooth, differentiable, monotonic trajectories at the target evaluation times $\hat t$ (see Algorithm \ref{alg:locreg-pchip-v}).

\begin{algorithm}
\caption{LOCREG-PCHIP-V Interpolation}
\label{alg:locreg-pchip-v}
\begin{algorithmic}
    \REQUIRE $[T,X,V] = [(t_1, x_1, v_1), (t_2, x_2, v_2), \ldots, (t_n, x_n, v_n)]$ where $t_1 < t_2 < \cdots < t_n$
    \REQUIRE $k_x, k_v$ \COMMENT{Neighborhood sizes for position and velocity}
    \STATE $f_{x} \leftarrow \text{LOCREG}(T,X, k_x)$  \COMMENT{Fit LOCREG model for locations with neighborhood $k_x$}
    \STATE $f_v \leftarrow \text{LOCREG}(T,V, k_v)$ \COMMENT{Fit LOCREG model for velocities with neighborhood $k_v$}
    \FOR{$i = 1, 2, \ldots, n$}
        \STATE $\tilde x_i \leftarrow f_x(t_i)$ \COMMENT{Predict locations $\tilde X$ and velocities $\tilde V$ at original data points}
        \STATE $\tilde v_i \leftarrow f_v(t_i)$ \COMMENT{Next, enforce monotonicity on smoothed positions $Y$}
        \IF{$\tilde x_i \geq \tilde x_{i-1}$ OR $i = 1$} 
            \STATE $y_i \leftarrow \tilde x_{i}$
        \ELSE
            \STATE $y_i \leftarrow y_{i-1}$
            \IF{$i < n$}
                \STATE $u_i \leftarrow \text{max} \left\{ \frac{y_{i+1} - y_{i-1}}{t_{i+1} - t_{i-1}}, 0 \right\}$ \COMMENT{Enforce monotonicity on smoothed velocities $U$}
            \ENDIF
        \ENDIF
    \ENDFOR
    \STATE $\hat f \leftarrow \text{VCHIP-ME}(T, Y, U)$
    \RETURN $\hat f$ \COMMENT{Return interpolation function for prediction at new time points $\hat t$}
\end{algorithmic}
\end{algorithm}

\subsubsection{Velocity-Aware Smoothing Spline (V-SPLINE)}
The V-SPLINE method represents a sophisticated approach to trajectory reconstruction that simultaneously incorporates both position and velocity observations while enforcing smoothness through regularization \citep{cao_v-spline_2021}.

The method constructs a cubic Hermite spline with knots at each observation time $t_i$, where both position $x_i$ and velocity $v_i$ values are specified. The spline parameters $\boldsymbol{\theta} = [\theta_1, \theta_2, \ldots, \theta_{2n}]^T$ represent alternating position and velocity values at each knot, where $\theta_{2i-1}$ corresponds to position at time $t_i$ and $\theta_{2i}$ corresponds to velocity at time $t_i$.

The objective function combines data fitting terms with a smoothness penalty:
\begin{align}
\min_{\boldsymbol{\theta}} \quad &\|\mathbf{B}\boldsymbol{\theta} - \mathbf{x}\|^2 + \gamma \|\mathbf{C}\boldsymbol{\theta} - \mathbf{v}\|^2 + n \boldsymbol{\theta}^T \boldsymbol{\Omega} \boldsymbol{\theta},
\end{align}

where:
\begin{itemize}
\item $\mathbf{B} \in \mathbb{R}^{n \times 2n}$ is the position observation matrix with $B_{i,2i-1} = 1$ and all other entries zero
\item $\mathbf{C} \in \mathbb{R}^{n \times 2n}$ is the velocity observation matrix with $C_{i,2i} = 1$ and all other entries zero
\item $\mathbf{x} = [x_1, x_2, \ldots, x_n]^T$ are the observed positions
\item $\mathbf{v} = [v_1, v_2, \ldots, v_n]^T$ are the observed velocities
\item $\gamma \geq 0$ is the velocity weight parameter controlling the relative importance of velocity observations
\item $\boldsymbol{\Omega} \in \mathbb{R}^{2n \times 2n}$ is the penalty matrix enforcing smoothness
\end{itemize}

The penalty matrix $\boldsymbol{\Omega}$ penalizes the integrated squared second derivative (curvature) of cubic Hermite polynomials, promoting smooth trajectories (a very common approach \citep{green_nonparametric_1993}). For each interval $[t_i, t_{i+1}]$, the penalty matrix entries are:
\begin{align}
    \Omega^{(i)}_{2i-1,2i-1} = \Omega^{(i)}_{2i+1,2i+1} &= \lambda_i \frac{12}{(t_{i+1} - t_i)^3}, \\
    \Omega^{(i)}_{2i-1,2i+1} = \Omega^{(i)}_{2i+1,2i-1} &= -\lambda_i \frac{12}{(t_{i+1} - t_i)^3},\\
    \Omega^{(i)}_{2i-1,2i} = \Omega^{(i)}_{2i,2i-1} = \Omega^{(i)}_{2i-1,2i+2} = \Omega^{(i)}_{2i+2,2i-1} &= \lambda_i \frac{6}{(t_{i+1} - t_i)^2}, \\
    \Omega^{(i)}_{2i,2i+1} = \Omega^{(i)}_{2i+1,2i} = \Omega^{(i)}_{2i+1,2i+2} = \Omega^{(i)}_{2i+2,2i+1} &= -\lambda_i \frac{6}{(t_{i+1} - t_i)^2}, \\
    \Omega^{(i)}_{2i,2i} = \Omega^{(i)}_{2i+2,2i+2} &= \lambda_i \frac{4}{t_{i+1} - t_i},\\
    \Omega^{(i)}_{2i,2i+2} = \Omega^{(i)}_{2i+2,2i} &= \lambda_i \frac{2}{t_{i+1} - t_i}.
    \end{align}

We consider the adaptive version of the V-SPLINE algorithm with penalty weights $\lambda_i$ that can vary across intervals. When using adaptive penalties with parameter $\eta > 0$, the weights are calculated as:
\begin{align}
\lambda_i = \eta \cdot \frac{t_{i+1} - t_i}{v_{avg,i}^2},
\end{align}
where $v_{avg,i} = (x_{i+1} - x_i)/(t_{i+1} - t_i)$ is the average velocity over interval $i$.

The optimization problem has a closed-form solution:
\begin{align}
\boldsymbol{\theta}^* = \left(\mathbf{B}^T\mathbf{B} + \gamma \mathbf{C}^T\mathbf{C} + n\boldsymbol{\Omega}\right)^{-1}\left(\mathbf{B}^T\mathbf{x} + \gamma \mathbf{C}^T\mathbf{v}\right).
\label{eq:v_obj}
\end{align}

For evaluation at arbitrary time points $\hat{t}$, the method uses cubic Hermite interpolation between adjacent knots:
\begin{align}
\hat{x}(\hat{t}) &= h_{00}(s)\theta_{2i-1} + h_{10}(s)(t_{i+1}-t_i)\theta_{2i} + h_{01}(s)\theta_{2i+1} + h_{11}(s)(t_{i+1}-t_i)\theta_{2i+2},\\
\hat{v}(\hat{t}) &= \dot{h}_{00}(s)\frac{\theta_{2i-1}}{t_{i+1}-t_i} + \dot{h}_{10}(s)\theta_{2i} + \dot{h}_{01}(s)\frac{\theta_{2i+1}}{t_{i+1}-t_i} + \dot{h}_{11}(s)\theta_{2i+2},
\end{align}
where the Hermite basis functions are defined as in the PCHIP method.

\subsubsection{Velocity-Aware Smoothing Spline with Monotonicity Penalization (V-SPLINE-MP)}
V-SPLINE does not guarantee monotonicity. V-SPLINE-MP addresses this limitation by incorporating monotonicity constraints both in the preprocessing of velocity observations and as a penalty in the objective function. While strict enforcement of this approach would require constrained optimization (which is computationally expensive \citep{kano_velocity_2017}), this approach can achieve near-monotone trajectories with appropriate parameter selection.

First, the method applies PCHIP-style constraints (equations \eqref{eq:fc1} through \eqref{eq:fcn}) to the observed velocity data to ensure local monotonicity consistency. 

Next, The key innovation of V-SPLINE-MP is the addition of a monotonicity penalty term to encourage velocity consistency with the monotonic direction of position changes. The complete optimization problem becomes:
\begin{align}
\min_{\boldsymbol{\theta}} \quad &\|\mathbf{B}\boldsymbol{\theta} - \mathbf{x}\|^2 + \gamma \|\mathbf{C}\boldsymbol{\theta} - \mathbf{u}\|^2 + n \boldsymbol{\theta}^T \boldsymbol{\Omega}_{smooth} \boldsymbol{\theta} + \boldsymbol{\theta}^T \boldsymbol{\Omega}_{mono} \boldsymbol{\theta} - 2\mathbf{b}_{mono}^T\boldsymbol{\theta},
\end{align}
where $\mathbf{u} = [u_1, u_2, \ldots, u_n]^T$ are the constrained velocities from the first step, $\boldsymbol{\Omega}_{smooth}$ is the original V-SPLINE smoothness penalty matrix, and $\boldsymbol{\Omega}_{mono}$ and $\mathbf{b}_{mono}$ encode the monotonicity penalty (quadratic and linear terms respectively).

For each interval $[t_i, t_{i+1}]$, the secant slope $s_i = (x_{i+1} - x_i)/(t_{i+1} - t_i)$ maintains a consistent monotonic direction due to the PCHIP-style constraints applied in Stage 1. We use this secant slope as the target velocity.

The monotonicity penalty encourages the velocity parameters $\theta_{2i}$ and $\theta_{2i+2}$ (velocities at interval endpoints $t_i$ and $t_{i+1}$) to approach this target:
\begin{align}
\frac{\mu}{t_{i+1} - t_i} \left[(\theta_{2i} - s_i)^2 + (\theta_{2i+2} - s_i)^2\right],
\end{align}
where $\mu > 0$ controls the penalty strength and the normalization ensures consistent weighting across intervals of different lengths.

Expanding the quadratic penalty terms across all intervals $i = 1, \ldots, n-1$, each interval $i$, contributes:
\begin{align}
\Omega^{(i)}_{mono,2i,2i} &= \frac{\mu}{t_{i+1} - t_i}, \\
\Omega^{(i)}_{mono,2i+2,2i+2} &= \frac{\mu}{t_{i+1} - t_i},\\
b^{(i)}_{mono,2i} &\mathrel{+}= \frac{\mu s_i}{t_{i+1} - t_i}, \\
b^{(i)}_{mono,2i+2} &\mathrel{+}= \frac{\mu s_i}{t_{i+1} - t_i}.
\end{align}
Note that the constant term $s_i^2$ does not affect the solution so it is omitted. 

Combining the smoothness and monotonicity penalties into $\boldsymbol{\Omega}_{total} = \boldsymbol{\Omega}_{smooth} + \boldsymbol{\Omega}_{mono}$, and setting the gradient to zero yields the closed-form solution:
\begin{align}
\boldsymbol{\theta}^* = \left(\mathbf{B}^T\mathbf{B} + \gamma \mathbf{C}^T\mathbf{C} + n\boldsymbol{\Omega}_{total}\right)^{-1}\left(\mathbf{B}^T\mathbf{x} + \gamma \mathbf{C}^T\mathbf{u} + \mathbf{b}_{mono}\right).
\end{align}

The monotonicity penalty weight $\mu$ allows practitioners to control the trade-off between strict monotonicity and data fidelity. 

\subsubsection{Velocity-Aware Smoothing Spline with Monotonicity Enforcement (V-SPLINE-ME)}
The V-SPLINE-ME approach first calculates the V-SPLINE on the original data (equation \eqref{eq:v_obj}), which acts as a smoothing step. Then, monotonicity is enforced on the smoothed points (following the same logic as the LOCREG-PCHIP-V, equations \eqref{eq:locrec_mon1} and \eqref{eq:locrec_mon2}), modifying both point locations and velocities to maintain consistency. 

Finally, we ensure monotonicity with equations \eqref{eq:fc1} through \eqref{eq:fcn} and apply standard cubic Hermite interpolation just like the VCHIP-ME and LOCREC-PCHIP-V approaches, where now the V-SPLINE approach has been used to perform a smoothing which better respects the physical connection between location and velocity. 

\subsection{Overview of Interpolation Approaches}
Based on their general characteristics (shown in Table \ref{tab:method_comparison}), the most promising algorithms are the LOCREG-PCHIP-V and the V-SPLINE-ME which are the only two methods which meet all of the evaluation criteria. However, tuning the parameters needed for these methods can be challenging, especially when generalizing across many trajectories. In addition to these methods, the VCHIP-ME method is particularly promising if data is already highly regular and does not need smoothing. In that case, the VCHIP-ME method does not need any input parameters and can be evaluated very quickly.

\begin{table}[h]
\centering
\caption{Overview of Trajectory Reconstruction Methods}
\label{tab:method_comparison}
\begin{tabular}{lcccccc}
\toprule
Algorithm & MON\textsuperscript{1} & CUB\textsuperscript{2} & DIFF\textsuperscript{3} & ERR\textsuperscript{4} & VEL\textsuperscript{5} & PARAMS\textsuperscript{6} \\
\midrule
LSEG & \checkmark & $\times$ & $\times$ & $\times$ & $\times$ & 0 \\
PCHIP & \checkmark & \checkmark & \checkmark & $\times$ & $\times$ & 0 \\
LOCREG & $\times$ & $\times$ & $\times$ & \checkmark & $\times$ & 1 \\
LOCREG-PCHIP & \checkmark & \checkmark & \checkmark & \checkmark & $\times$ & 1 \\
LVMI & $\times$ & $\times$ & $\times$ & $\times$ & \checkmark & 0 \\
VCHIP & $\times$ & \checkmark & \checkmark & $\times$ & \checkmark & 0 \\
VCHIP-ME & \checkmark & \checkmark & \checkmark & $\times$ & \checkmark & 0 \\
PCHIP-VCHIP & \checkmark & \checkmark & \checkmark & $\times$ & \checkmark & 1 \\
LOCREG-V & $\times$ & $\times$ & $\times$ & \checkmark & \checkmark & 2 \\
LOCREG-PCHIP-V & \checkmark & \checkmark & \checkmark & \checkmark & \checkmark & 2 \\
V-SPLINE & $\times$ & \checkmark & \checkmark & \checkmark & \checkmark & 2 \\
V-SPLINE-MP & $\sim$ & \checkmark & \checkmark & \checkmark & \checkmark & 3 \\
V-SPLINE-ME & \checkmark & \checkmark & \checkmark & \checkmark & \checkmark & 2 \\
\bottomrule
\end{tabular}
\begin{tablenotes}
\footnotesize
\item\textsuperscript{1} MON: The trajectory is non-decreasing (monotonic)
\item\textsuperscript{2} CUB: The trajectory is made up of cubic polynomials
\item\textsuperscript{3} DIFF: The trajectory is once differentiable
\item\textsuperscript{4} ERR: Minimizes measurement error through smoothing
\item\textsuperscript{5} VEL: Uses velocity data in reconstruction
\item\textsuperscript{6} PARAMS: Number of tunable parameters
\end{tablenotes}
\end{table}

\section{Case Study}
\label{sec:case_study}
This section describes the performance of these thirteen methods using field data from Austin, TX, obtained from the 801 and 803 routes. These are high-frequency routes passing through downtown, and are prone to data gaps, particularly around tall buildings. We evaluated 7,620 complete trajectories from September, 2024.

We evaluated these approaches using a sparse dataset (2,534,202 AVL records, on average 332 per trajectory) and a dense dataset (6,994,372 records, averaging 918 per trajectory). The average time gap between data points is 16.49 seconds for the sparse dataset and 5.96 for the dense dataset; the average distance gaps are 294.9 ft and 106.6 ft, respectively. Comparing these datasets is important to understand the storage and performance tradeoffs when working with AVL data. The next subsections explain our preprocessing procedure and evaluation framework.

\subsection{Preprocessing}
To ensure a fair comparison between the two datasets, we consider only complete trajectories present in both datasets. The raw AVL data was processed in five steps: 

First, we matched AVL with APC data to consider only passenger serving trips (eliminating training buses and incomplete routes). Second, we matched individual data points to trips, provided that the bearing was within 20 degrees and the distance to the route did not exceed 200 feet. Third, we implemented a complete trip filter to eliminate trips with data gaps of more than 10 minutes or 1 mile. 

The fourth and largest data processing step was an outlier filter, which removed large forward or backward jumps exceeding 500 ft, which would imply a velocity greater than 45 miles per hour. We also removed all backward jumps exceeding 200 ft regardless of the implied velocity, and adjusted point locations of those points where backtracking was less than 200 ft by moving them forward until a monotonic trajectory was achieved. Finally, we eliminated the starting and ending points of the trips when the vehicle was out of service or when it was stopped at the beginning or end of the route. 

Together, these adjustments create the necessary $(t, x, v)$ points that serve as input for the interpolation approaches. 

\subsection{Evaluation Framework}
We evaluated trajectory reconstruction performance using three distinct tests: a data-driven approach, a physics-based approach, and a realistic implementation-driven approach. Together, these evaluations offer insight into the performance of each method both at the large-scale and for more fine-grained analysis. 

The first test evaluated fit of the trajectory by $5\%$ of the data points from each trajectory and comparing estimated versus recorded location and velocity at those points. Evaluation criteria included the root mean squared error (RMSE) and mean absolute error (MAE) for both location (distance along the route) and velocity. 

Next, we validated the velocity and acceleration profiles based on realistic acceleration bounds and known stop information. The AVL data identifies when a vehicle is ``stopped at'' a bus stop. We determined the percentage of the time the vehicle is listed as stopped and also stopped in the interpolation. We examined acceleration bounds (tight: $-5.79$ to 4.26 ft/s$^2$, loose: $-7.77$ to 5.43 ft/s$^2$) and stop detection at 2, 5, and 10 ft/s thresholds. Note that inconsistencies exist in the original vehicle record -- $20\%$ of the time vehicles are listed as stopped, they are recorded traveling at velocities greater than 5 ft/s. Given these inconsistencies, no interpolation approach can completely eliminate errors, though they should be minimized where possible. 

The final test assessed performance for practical application. Since AVL data is often used for localized assessments near stops or signals, we calculated metrics 300 ft upstream of each signalized intersection. We calculated average travel time, average speed, average speed volatility, and average deceleration rate. Then, using one model selected as the baseline from the dense dataset, we calculated percent error for the other metrics on both datasets to compare the value of each model. 

These three tests allow us to analyze which model best matches raw data, reflects physical characteristics of typical bus trips, and is most suitable for evaluating real-world traffic phenomena.

\section{Results}
\label{sec:results}
According to our evaluation framework, the results are organized into three sections: overall trajectory fit and computational analysis. physical trajectory validation, and real-world value analysis. 

\subsection{Trajectory Fit}
We evaluate the interpolation methods using RMSE and MAE of position and velocity, average portion of trajectory that violates monotonicity, proportion of trajectories with at least one monotonicity violation, and computation time. Table \ref{tab:performance_comparison_combined} presents the results for both sparse and dense datasets. The findings clearly demonstrate the advantages of velocity-aware approaches and high data density. 
 
Position-only methods (LSEG, PCHIP, LOCREG, and LOCREG-PCHIP) exhibit fundamentally limited performance due to their reliance solely on spatial coordinates. Linear segmentation (LSEG) performs the poorest, achieving position RMSE values of 117.12 ft (sparse) and 32.42 ft (dense), reflecting the inadequacy of linear interpolation in capturing the complex acceleration patterns. PCHIP improves performance with RMSE of 89.39 ft (sparse) and 19.73 ft (dense) but cannot incorporate velocity information, potentially violating realistic acceleration bounds despite smooth profiles. Local regression approaches (LOCREG and LOCREG-PCHIP) both outperform LSEG but underperform PCHIP approach, suggesting some oversmoothing. While \citet{huang_reconstructing_2023} found that smoothing was beneficial, it does not provide the same advantages for our data. This may indicate that there were relatively few outliers in our dataset, so the smoothing may be counterproductive in some cases when rapidly oscillating velocities accurately reflect the vehicle's behavior in congested traffic or near signals.

Velocity-aware methods demonstrate superior performance. VCHIP-ME emerges as one of the best performers, achieving position RMSE values of 61.55 ft (sparse) and 14.55 ft (dense), improvements of $31\%$ and $26\%$ over PCHIP. Velocity RMSE values were similarly improved on by $15\%$ and $8.5\%$. The V-SPLINE variants perform comparably; V-SPLINE achieves 58.33 ft RMSE (sparse) and 14.34 ft RMSE (dense), though at substantially higher computational cost. Regarding the LOCREG and V-SPLINE methods, the smoothing element of these approaches does not provide substantial benefits on these datasets. Though V-SPLINE is the best performer on the sparse dataset, it is only marginally better than the simpler VCHIP and VCHIP-ME methods, and actually does worse than those methods on the dense dataset. The superior performance of velocity-aware methods stems from their ability to capture the true dynamics of bus operations, including acceleration, deceleration, and stop-and-go patterns. 

Monotonicity enforcement proves critical despite accuracy trade-offs.  Unconstrained methods including VCHIP and V-SPLINE, achieve marginally better position accuracy (58.15 ft for VCHIP vs 61.55 ft RMSE for VCHIP-ME in sparse data). However, they exhibit poor monotonic success rates of only $52\%$ for VCHIP and $58\%$ V-SPLINE in the sparse dataset. Conversely, monotonicity-enforced variants achieve $100\%$ success rates at approximately $5.5\%$ position accuracy cost. This trade-off is particularly pronounced for V-SPLINE-MP, which shows substantial performance degradation (73.35 ft RMSE sparse), suggesting excessive smoothing compromises the ability to capture rapid velocity variations.

Data density profoundly impacts reconstruction accuracy. Dense datasets (5.96-second intervals) provide 4-6 times better position MAE and 2-3 times better velocity MAE than sparse datasets (16.49-second intervals). The gains are actually the highest among the best performing models, suggesting that the combination of the best models with the best data can provide a powerful boost to predictive power. VCHIP-ME demonstrates this effect with $76\%$ position RMSE reduction (61.55 ft to 14.55 ft) and $50\%$ velocity RMSE reduction (5.16 ft/s to 2.58 ft/s) between sparse and dense datasets. These results underscore that investing in increased data collection frequency yields substantial returns in reconstruction quality.

Computational efficiency varies significantly across methods. PCHIP and VCHIP-based methods require 0.43-1.66 milliseconds for sparse reconstruction, while V-SPLINE approaches demand 240-245 milliseconds—nearly two orders of magnitude higher due to $O(n^3)$ versus $O(n)$ complexity. Methods with local smoothing also take somewhat longer than those without, and this time can vary depending on the choice of $k$. However, with fixed $k$, the computation time still grows at order $O(n)$. In dense datasets, most methods achieve sub-second computation times except V-SPLINE variants (5.24-5.26 seconds), potentially limiting real-time applications. 

Overall, VCHIP-ME provides an optimal balance of accuracy and efficiency, delivering top-tier reconstruction quality with minimal computational overhead, making it suitable for both offline analysis and real-time applications. We also find that the substantial performance gains from dense data collection justify an increased investment in higher-frequency AVL data acquisition systems. To be clear, the data set for the case study consists of complex trajectories through a congested downtown area with hundreds of signalized intersections. For applications where trajectory data contains less inherent variability, the approaches with smoothing may provide more benefits and the need for data density might be less critical. 

\begin{landscape}
\begin{table}[H]
\centering
\fontsize{11}{12}\selectfont
\setlength{\tabcolsep}{3pt}
\caption{Performance Comparison of Trajectory Reconstruction Methods for Sparse and Dense Datasets}
\label{tab:performance_comparison_combined}
\begin{threeparttable}
\begin{tabular}{lcccccccccccccc}
\toprule
& \multicolumn{7}{c}{\textbf{Sparse Dataset}} & \multicolumn{7}{c}{\textbf{Dense Dataset}} \\
\cmidrule(lr){2-8} \cmidrule(lr){9-15}
Algorithm & \makecell{RMSE\\ Pos} & \makecell{RMSE\\ Vel} & \makecell{MAE\\ Pos} & \makecell{MAE\\ Vel} & \makecell{Viol\\ Rate} & \makecell{Mon\\ Success} & \makecell{Comp\\ Time} & \makecell{RMSE\\ Pos} & \makecell{RMSE\\ Vel} & \makecell{MAE\\ Pos} & \makecell{MAE\\ Vel} & \makecell{Viol\\ Rate} & \makecell{Mon\\ Success} & \makecell{Comp\\ Time} \\
\midrule
LSEG & \makecell{117.12\\(39.53)} & \makecell{19.60\\(3.15)} & \makecell{80.95\\(31.10)} & \makecell{14.08\\(2.96)} & 0.00 & 1.00 & 0.56 & \makecell{32.42\\(50.90)} & \makecell{12.07\\(4.13)} & \makecell{19.28\\(38.06)} & \makecell{7.53\\(1.78)} & 0.00 & 1.00 & 0.65 \\
PCHIP & \makecell{89.39\\(38.87)} & \makecell{6.04\\(1.67)} & \makecell{59.91\\(29.70)} & \makecell{4.07\\(1.26)} & 0.00 & 1.00 & 0.43 & \makecell{19.73\\(52.66)} & \makecell{2.82\\(4.39)} & \makecell{10.88\\(38.05)} & \makecell{1.36\\(1.12)} & 0.00 & 0.99 & 0.54 \\
LOCREG & \makecell{105.28\\(106.95)} & \makecell{6.70\\(2.44)} & \makecell{69.98\\(53.43)} & \makecell{4.58\\(1.53)} & 0.01 & 0.74 & 4.54 & \makecell{29.81\\(119.29)} & \makecell{3.86\\(15.04)} & \makecell{13.75\\(59.94)} & \makecell{1.51\\(2.00)} & 0.02 & 0.17 & 9.04 \\
LOCREG-PCHIP & \makecell{97.84\\(39.41)} & \makecell{6.43\\(1.67)} & \makecell{66.88\\(30.01)} & \makecell{4.37\\(1.28)} & 0.00 & 1.00 & 4.22 & \makecell{21.64\\(53.36)} & \makecell{2.88\\(4.23)} & \makecell{12.46\\(38.91)} & \makecell{1.44\\(1.11)} & 0.00 & 0.99 & 8.60 \\
LVMI & \makecell{124.79\\(60.23)} & \makecell{15.26\\(2.92)} & \makecell{77.57\\(40.35)} & \makecell{10.18\\(2.51)} & 0.02 & 0.59 & 0.43 & \makecell{31.82\\(76.86)} & \makecell{7.92\\(1.61)} & \makecell{18.64\\(49.41)} & \makecell{4.90\\(1.36)} & 0.04 & 0.06 & 0.85 \\
VCHIP & \makecell{58.15\\(34.18)} & \makecell{4.96\\(1.62)} & \makecell{38.44\\(24.16)} & \makecell{3.30\\(1.17)} & 0.02 & 0.52 & 0.45 & \makecell{14.17\\(48.59)} & \makecell{2.47\\(4.59)} & \makecell{7.46\\(33.34)} & \makecell{1.12\\(1.16)} & 0.04 & 0.07 & 0.88 \\
VCHIP-ME & \makecell{61.55\\(35.53)} & \makecell{5.16\\(1.61)} & \makecell{40.02\\(25.55)} & \makecell{3.47\\(1.16)} & 0.00 & 1.00 & 0.83 & \makecell{14.55\\(50.35)} & \makecell{2.58\\(4.61)} & \makecell{7.44\\(34.49)} & \makecell{1.17\\(1.15)} & 0.00 & 1.00 & 1.84 \\
PCHIP-VCHIP & \makecell{72.56\\(35.77)} & \makecell{5.48\\(1.62)} & \makecell{48.17\\(26.63)} & \makecell{3.70\\(1.19)} & 0.00 & 1.00 & 1.66 & \makecell{16.55\\(49.65)} & \makecell{2.65\\(4.50)} & \makecell{8.81\\(34.96)} & \makecell{1.24\\(1.13)} & 0.00 & 1.00 & 2.83 \\
LOCREG-V & \makecell{100.85\\(58.47)} & \makecell{12.77\\(2.50)} & \makecell{69.47\\(36.87)} & \makecell{9.01\\(2.11)} & 0.01 & 0.66 & 7.08 & \makecell{27.38\\(98.24)} & \makecell{4.91\\(1.58)} & \makecell{13.14\\(47.42)} & \makecell{3.09\\(1.30)} & 0.03 & 0.11 & 16.19 \\
LOCREG-PCHIP-V & \makecell{71.08\\(36.33)} & \makecell{5.54\\(1.62)} & \makecell{47.72\\(26.12)} & \makecell{3.77\\(1.19)} & 0.00 & 1.00 & 10.27 & \makecell{16.53\\(51.24)} & \makecell{2.67\\(4.51)} & \makecell{9.04\\(35.95)} & \makecell{1.26\\(1.14)} & 0.00 & 1.00 & 23.02 \\
V-SPLINE & \makecell{58.33\\(34.23)} & \makecell{4.96\\(1.63)} & \makecell{38.00\\(24.28)} & \makecell{3.26\\(1.17)} & 0.03 & 0.58 & 240.21 & \makecell{14.34\\(48.51)} & \makecell{2.59\\(4.59)} & \makecell{7.66\\(33.33)} & \makecell{1.23\\(1.15)} & 0.06 & 0.18 & 5247.75 \\
V-SPLINE-MP & \makecell{73.35\\(33.93)} & \makecell{6.36\\(1.47)} & \makecell{49.69\\(24.96)} & \makecell{4.59\\(1.10)} & 0.01 & 0.86 & 244.03 & \makecell{27.84\\(49.06)} & \makecell{10.32\\(4.00)} & \makecell{16.36\\(33.99)} & \makecell{7.06\\(1.29)} & 0.01 & 0.85 & 5255.09 \\
V-SPLINE-ME & \makecell{60.69\\(35.43)} & \makecell{5.14\\(1.60)} & \makecell{39.70\\(25.49)} & \makecell{3.47\\(1.16)} & 0.00 & 1.00 & 245.24 & \makecell{14.74\\(50.30)} & \makecell{2.62\\(4.59)} & \makecell{7.90\\(34.43)} & \makecell{1.24\\(1.14)} & 0.00 & 1.00 & 5266.15 \\
\bottomrule
\end{tabular}
\begin{tablenotes}
\footnotesize
\item Values shown as mean with standard deviation in parentheses.
\item RMSE: root mean square error; MAE: mean absolute error; Pos: position; Vel: velocity
\item Viol Rate: mean violation rate (monotonicity violations per trajectory)
\item Mon Success: monotonic success rate (proportion of trajectories that are completely monotonic)
\item Comp Time: mean computation time (milliseconds)
\end{tablenotes}
\end{threeparttable}
\end{table}
\end{landscape}

\subsection{Velocity and Acceleration Profile Validation}
We evaluate physical realism by examining acceleration patterns and stop behavior characteristics. Table \ref{tab:profile_comparison_combined} presents acceleration bounds adherence and stop detection across velocity thresholds, providing insights into how well each method captures the underlying physics of bus movement.

Acceleration constraint adherence reveals substantial differences among  methods. LVMI achieves perfect adherence ($100.00\%$) since the acceleration at all estimation points is 0 and it is undefined between them, though this comes at the cost of substantially higher position and velocity errors. Position-only methods show mixed performance, with PCHIP achieving $98.28\%$ adherence to the tighter bounds in sparse data and 96.64\% in dense data, while LSEG performs notably worse at $96.61\%$ (sparse) and $93.90\%$ (dense). The poor acceleration adherence of the LOCREG approach reflects its inability to capture smooth acceleration transitions. That method has no monotonicity guarantees and the acceleration and velocity profiles are more artificial since the smoothing works on the position data directly without consideration of physically realistic velocity curves. 

Velocity-aware methods demonstrate superior acceleration constraint adherence while maintaining high trajectory accuracy. VCHIP-ME achieves $98.72\%$ (sparse) and $97.21\%$ (dense) adherence for tighter bounds, representing an optimal balance between physical realism and reconstruction accuracy. The unconstrained VCHIP method performs slightly better at $99.51\%$ (sparse) and $97.68\%$ (dense) for the tighter bounds, but violates monotonicity. Notably,  V-SPLINE variants show excellent adherence, with V-SPLINE achieving $99.36\%$ (sparse) and $97.90\%$ (dense), demonstrating effective acceleration constraint incorporation. However, V-SPLINE-MP performs poorly in dense datasets, suggesting monotonicity constraints create instabilities in highly constrained trajectories.

Stop detection analysis across velocity thresholds (2, 5, and 10 ft/s) reveals how effectively each method captures stationary periods. Higher thresholds naturally increase detection rates. VCHIP-ME demonstrates consistent performance: $83.37\%$ (2 ft/s), $89.73\%$ (5 ft/s), and $94.78\%$ (10 ft/s) in sparse data, with comparable dense data performance. This progression reflects realistic deceleration and acceleration patterns around stops.

Acceleration adherence is generally higher in sparse datasets while stop detection improves in dense datasets. This pattern is indicative of the instabilities that can arise when collecting dense data that contains large errors. For top methods, 97-$99\%$ acceleration adherence is excellent, though denser data requires proportionally reduced errors to maintain performance. Stop detection differences between sparse and dense datasets are modest, suggesting fundamental stop-and-go patterns are captured adequately even at lower temporal resolutions.

Overall, VCHIP-ME and V-SPLINE-ME have good acceleration adherence rates and stop detection performance, demonstrating that this method successfully balances multiple physical constraints. The substantial degradation in acceleration adherence observed for certain methods (particularly LOCREG-V and V-SPLINE-MP) highlights the importance of careful constraint implementation, as poorly designed velocity or monotonicity constraints can introduce artifacts that violate other physical properties. These results reinforce the conclusion that VCHIP-ME provides the most practical approach for AVL trajectory reconstruction, delivering physically plausible trajectories that respect both kinematic constraints and operational characteristics of bus systems.

\begin{landscape}
\begin{table}[htbp]
\centering
\fontsize{11}{12}\selectfont
\caption{Profile Validation of Trajectory Reconstruction Methods for Sparse and Dense Datasets}
\label{tab:profile_comparison_combined}
\begin{threeparttable}
\begin{tabular}{lcccccccccc}
\toprule
& \multicolumn{5}{c}{\textbf{Sparse Dataset}} & \multicolumn{5}{c}{\textbf{Dense Dataset}} \\
\cmidrule(lr){2-6} \cmidrule(lr){7-11}
Algorithm & \makecell{Tight\\Accel} & \makecell{Loose\\Accel} & \makecell{Stops\\2 ft/s} & \makecell{Stops\\5 ft/s} & \makecell{Stops\\10 ft/s} & \makecell{Tight\\Accel} & \makecell{Loose\\Accel} & \makecell{Stops\\2 ft/s} & \makecell{Stops\\5 ft/s} & \makecell{Stops\\10 ft/s} \\
\midrule
LSEG & \makecell{96.61\\(0.52)} & \makecell{96.93\\(0.48)} & \makecell{70.73\\(13.64)} & \makecell{75.30\\(12.10)} & \makecell{80.75\\(10.44)} & \makecell{93.90\\(1.01)} & \makecell{94.84\\(0.86)} & \makecell{75.58\\(9.57)} & \makecell{81.12\\(9.10)} & \makecell{87.40\\(8.60)} \\
PCHIP & \makecell{98.28\\(0.60)} & \makecell{99.31\\(0.30)} & \makecell{81.98\\(9.42)} & \makecell{89.17\\(6.90)} & \makecell{94.95\\(5.28)} & \makecell{96.64\\(0.97)} & \makecell{98.47\\(0.52)} & \makecell{83.04\\(8.14)} & \makecell{88.34\\(7.60)} & \makecell{93.48\\(7.40)} \\
LOCREG & \makecell{99.28\\(1.09)} & \makecell{99.60\\(0.86)} & \makecell{74.96\\(11.07)} & \makecell{83.64\\(8.36)} & \makecell{91.83\\(6.13)} & \makecell{97.50\\(2.05)} & \makecell{98.68\\(1.69)} & \makecell{77.94\\(8.69)} & \makecell{86.01\\(7.81)} & \makecell{93.12\\(7.45)} \\
LOCREG-PCHIP & \makecell{99.17\\(0.37)} & \makecell{99.71\\(0.16)} & \makecell{77.49\\(10.91)} & \makecell{86.40\\(7.75)} & \makecell{93.81\\(5.48)} & \makecell{97.33\\(0.84)} & \makecell{98.89\\(0.42)} & \makecell{81.54\\(8.55)} & \makecell{87.38\\(7.78)} & \makecell{93.15\\(7.43)} \\
LVMI & \makecell{100.00\\(0.00)} & \makecell{100.00\\(0.00)} & \makecell{82.36\\(10.60)} & \makecell{84.17\\(9.82)} & \makecell{87.55\\(8.71)} & \makecell{100.00\\(0.00)} & \makecell{100.00\\(0.00)} & \makecell{82.44\\(8.98)} & \makecell{85.13\\(8.57)} & \makecell{90.31\\(8.10)} \\
VCHIP & \makecell{99.51\\(0.32)} & \makecell{99.93\\(0.09)} & \makecell{83.14\\(8.56)} & \makecell{87.96\\(7.03)} & \makecell{93.02\\(5.60)} & \makecell{97.68\\(1.10)} & \makecell{99.28\\(0.56)} & \makecell{82.10\\(8.03)} & \makecell{87.20\\(7.63)} & \makecell{92.85\\(7.40)} \\
VCHIP-ME & \makecell{98.72\\(0.50)} & \makecell{99.56\\(0.22)} & \makecell{83.37\\(8.84)} & \makecell{89.73\\(6.60)} & \makecell{94.78\\(5.28)} & \makecell{97.21\\(1.02)} & \makecell{98.88\\(0.54)} & \makecell{83.26\\(8.11)} & \makecell{88.44\\(7.58)} & \makecell{93.38\\(7.40)} \\
PCHIP-VCHIP & \makecell{98.66\\(0.50)} & \makecell{99.50\\(0.24)} & \makecell{82.48\\(9.22)} & \makecell{89.55\\(6.74)} & \makecell{94.89\\(5.26)} & \makecell{97.25\\(0.90)} & \makecell{98.85\\(0.46)} & \makecell{83.09\\(8.14)} & \makecell{88.38\\(7.59)} & \makecell{93.43\\(7.40)} \\
LOCREG-V & \makecell{35.93\\(7.49)} & \makecell{37.62\\(7.47)} & \makecell{74.45\\(12.54)} & \makecell{80.85\\(10.52)} & \makecell{87.82\\(8.57)} & \makecell{40.05\\(7.69)} & \makecell{41.25\\(7.67)} & \makecell{77.22\\(9.49)} & \makecell{84.23\\(8.76)} & \makecell{91.68\\(8.20)} \\
LOCREG-PCHIP-V & \makecell{99.38\\(0.30)} & \makecell{99.81\\(0.13)} & \makecell{78.54\\(10.50)} & \makecell{87.42\\(7.32)} & \makecell{93.83\\(5.47)} & \makecell{97.80\\(0.91)} & \makecell{99.16\\(0.45)} & \makecell{81.64\\(8.61)} & \makecell{87.41\\(7.79)} & \makecell{93.13\\(7.42)} \\
V-SPLINE & \makecell{99.36\\(0.34)} & \makecell{99.88\\(0.11)} & \makecell{83.08\\(8.75)} & \makecell{88.81\\(6.85)} & \makecell{93.68\\(5.47)} & \makecell{97.90\\(0.95)} & \makecell{99.34\\(0.47)} & \makecell{81.38\\(8.19)} & \makecell{87.62\\(7.64)} & \makecell{93.37\\(7.40)} \\
V-SPLINE-MP & \makecell{91.12\\(2.64)} & \makecell{95.01\\(1.74)} & \makecell{81.65\\(9.52)} & \makecell{89.29\\(6.82)} & \makecell{95.37\\(5.27)} & \makecell{59.81\\(7.28)} & \makecell{63.83\\(6.47)} & \makecell{84.94\\(8.24)} & \makecell{91.46\\(7.58)} & \makecell{96.17\\(7.37)} \\
V-SPLINE-ME & \makecell{99.03\\(0.42)} & \makecell{99.73\\(0.17)} & \makecell{81.75\\(9.36)} & \makecell{88.85\\(6.79)} & \makecell{94.58\\(5.30)} & \makecell{97.88\\(0.94)} & \makecell{99.32\\(0.46)} & \makecell{81.40\\(8.33)} & \makecell{87.70\\(7.61)} & \makecell{93.42\\(7.38)} \\
\bottomrule
\end{tabular}
\begin{tablenotes}
\footnotesize
\item Values shown as mean percent success rate within each trajectory, with standard deviation in parentheses
\item Tight Accel: average percent of the time acceleration falls within bounds ($-5.79$, 4.26)
\item Loose Accel: average percent of the time acceleration falls within bounds ($-7.77$, 5.43)
\item Stops 2/5/10: average percent of the time vehicle is considered ``stopped'' at 2/5/10 ft/s threshold
\end{tablenotes}
\end{threeparttable}
\end{table}
\end{landscape}

\subsection{Implications for Practical Applications}
Table \ref{tab:intersection_performance_combined} provides operationally relevant insights by evaluating the performance of various methods at individual intersections where buses experience complex operational dynamics. Using VCHIP-ME as the baseline, this analysis calculates mean and mean absolute percentage error (MAPE) of performance metrics aggregated across all intersections. 

Velocity-aware methods maintain superior performance at intersections. VCHIP-ME achieves travel time MAPE of $5.63\%$ (sparse) and $0.00\%$ (dense), with a performance comparable to V-SPLINE-ME at $5.61\%$ (sparse) and $0.52\%$ (dense). Unconstrained VCHIP performs slightly better in sparse data ($5.26\%$ MAPE), but the minimal difference reinforces that monotonicity enforcement provides operational reliability without substantial accuracy penalties. The dramatic improvement in dense data performance across top-performing methods indicates that collecting higher temporal resolution data yields benefits for intersection-level analysis, where precise timing of acceleration and deceleration events is critical for operational metrics.

Speed volatility analysis reveals substantial trajectory smoothness variations. LSEG exhibits extremely high speed volatility with a MAPE of $186.64\%$ (sparse) and $107.11\%$ (dense), reflecting the limitations of linear interpolation. V-SPLINE-MP demonstrates severe instability with $218.67\%$ (sparse) and $420.88\%$ (dense) speed volatility MAPE, indicating that the monotonicity preservation approach introduces substantial artifacts that compromise trajectory realism. In contrast, VCHIP-ME maintains a reasonable speed volatility MAPE of $44.16\%$ (sparse) and $0.00\%$ (dense), demonstrating that velocity-aware methods with appropriate monotonicity enforcement can maintain smooth trajectories while preserving operational accuracy. 

VCHIP, VHIP-ME, and V-SPLINE-ME emerge as the top performers across both datasets. Speed volatility and declaration results reveal large errors between the baseline and other methods for the sparse dataset. Though still sensitive, the results in the dense dataset suggest that the best performing methods have errors of less than $1\%$ for travel time and speed, $5\%$ for speed volatility, and $25\%$ for deceleration, representing substantial improvement over sparse data.

\begin{table}[H]
\centering
\caption{Intersection-Level Performance Metrics for Sparse and Dense Datasets}
\label{tab:intersection_performance_combined}
\begin{threeparttable}
\begin{tabular}{lrrrrrrrr}
\toprule
\multirow{2}{*}{Algorithm} & \multicolumn{2}{c}{Travel Time} & \multicolumn{2}{c}{Speed} & \multicolumn{2}{c}{Speed Volatility} & \multicolumn{2}{c}{Deceleration} \\
\cmidrule(lr){2-3} \cmidrule(lr){4-5} \cmidrule(lr){6-7} \cmidrule(lr){8-9}
 & Mean & MAPE & Mean & MAPE & Mean & MAPE & Mean & MAPE \\
\midrule
&\multicolumn{8}{c}{Sparse Dataset} \\
\cmidrule(lr){2-9} 
LSEG & 19.02 & 13.38 & 27.45 & 42.79 & 0.35 & 186.64 & 3.55 & 223.25 \\
PCHIP & 23.46 & 9.68 & 27.21 & 9.23 & 0.50 & 71.84 & 0.88 & 76.71 \\
LOCREG & 25.90 & 21.22 & 27.14 & 10.57 & 0.51 & 71.15 & 0.67 & 74.93 \\
LOCREG-PCHIP & 23.10 & 12.04 & 26.99 & 11.13 & 0.44 & 75.72 & 0.71 & 78.88 \\
LVMI & 26.13 & 17.22 & 27.71 & 18.69 & 0.60 & 85.14 & 0.00 & 100.00 \\
VCHIP & 23.93 & 5.26 & 27.08 & 4.91 & 0.55 & 39.23 & 0.92 & 56.67 \\
VCHIP-ME & 23.76 & 5.63 & 27.16 & 5.36 & 0.53 & 44.16 & 0.97 & 58.39 \\
PCHIP-VCHIP & 23.59 & 7.40 & 27.14 & 7.00 & 0.50 & 55.66 & 0.86 & 65.72 \\
LOCREG-V & 26.68 & 38.12 & 25.85 & 19.62 & 0.49 & 72.73 & 0.07 & 100.23 \\
LOCREG-PCHIP-V & 23.38 & 8.51 & 26.96 & 7.63 & 0.47 & 53.69 & 0.82 & 65.25 \\
V-SPLINE & 23.75 & 5.17 & 27.07 & 4.89 & 0.53 & 40.84 & 0.87 & 59.18 \\
V-SPLINE-MP & 24.18 & 11.47 & 26.46 & 10.50 & 0.57 & 218.67 & 1.77 & 205.76 \\
V-SPLINE-ME & 23.70 & 5.61 & 27.08 & 5.28 & 0.51 & 45.02 & 0.91 & 59.82 \\ 
\cmidrule(lr){2-9} 
&\multicolumn{8}{c}{Dense Dataset} \\
\cmidrule(lr){2-9} 
LSEG & 23.67 & 2.96 & 27.08 & 21.81 & 0.62 & 107.11 & 5.23 & 259.47 \\
PCHIP & 24.01 & 1.20 & 26.97 & 1.48 & 0.61 & 21.10 & 1.44 & 37.33 \\
LOCREG & 26.74 & 13.38 & 27.00 & 2.24 & 0.64 & 30.86 & 1.04 & 52.00 \\
LOCREG-PCHIP & 23.96 & 1.72 & 26.97 & 1.94 & 0.59 & 24.82 & 1.23 & 42.68 \\
LVMI & 25.40 & 8.91 & 27.09 & 6.14 & 0.61 & 38.08 & 0.00 & 100.00 \\
VCHIP & 24.13 & 0.67 & 27.04 & 0.73 & 0.61 & 1.57 & 1.28 & 12.28 \\
VCHIP-ME & 24.03 & 0.00 & 27.01 & 0.00 & 0.61 & 0.00 & 1.50 & 0.00 \\
PCHIP-VCHIP & 24.02 & 0.63 & 26.98 & 0.80 & 0.61 & 11.47 & 1.41 & 24.54 \\
LOCREG-V & 28.78 & 28.60 & 26.33 & 6.16 & 0.59 & 33.30 & 0.06 & 101.80 \\
LOCREG-PCHIP-V & 23.98 & 0.81 & 27.00 & 0.99 & 0.60 & 10.41 & 1.30 & 21.58 \\
V-SPLINE & 23.99 & 0.64 & 27.01 & 0.79 & 0.60 & 4.37 & 1.12 & 21.21 \\
V-SPLINE-MP & 29.81 & 4.93 & 21.54 & 5.73 & 1.05 & 420.88 & 7.29 & 518.52 \\
V-SPLINE-ME & 23.99 & 0.52 & 26.99 & 0.66 & 0.59 & 3.94 & 1.23 & 17.03 \\
\bottomrule
\end{tabular}
\begin{tablenotes}
\footnotesize
\item Travel Time in seconds; Speed in ft/s; Speed Volatility; Deceleration in ft/s$^2$
\item MAPE: mean absolute percentage error  
\item Results aggregated across all intersections using VCHIP-ME as baseline
\end{tablenotes}
\end{threeparttable}
\end{table}

\newpage
\section{Conclusions}
\label{sec:conclusion}
This paper studied 13 trajectory interpolation methods for AVL data, including several novel velocity-aware approaches. Our evaluation framework examines four critical factors -- velocity, position, smoothing, and data density -- using mathematical error metrics, physical realism assessments (matching velocity and acceleration profiles to known stopped states), and practical operational metrics like deceleration rates and speed variability at intersection level. This is especially important since AVL data is generally used in very small spatial and temporal regimes (as larger areas can be studied in aggregate with APC data which is easier to process). We found robust evidence supporting the superiority of velocity-aware methods over position-only approaches and the important impact of data density on reconstruction accuracy. Our findings also revealed that, contrary to some existing literature, smoothing was generally unhelpful or even detrimental, potentially obscuring true variations in trajectories, though the V-SPLINE and V-SPLINE-ME methods showed promise where smoothing might be beneficial.

Among the methods tested, VCHIP-ME offered the best balance between high accuracy and computational efficiency. Its minimal overhead makes it suitable for both historical analysis and real-time applications, offering substantial gains in predictive power when combined with dense datasets. VCHIP-ME also ensures monotonicity without sacrificing the quality of the reconstruction and requires no parameter tuning. V-SPLINE-ME -- our novel V-SPLINE variant -- offered valuable insights and competitive performance in some aspects, though the substantially higher computational costs limits practical applicability in many scenarios. 

Based on these findings, we recommend the use of VCHIP-ME or V-SPLINE-ME for trajectory interpolation, with the choice depending on the certainty and characteristics of the original data. Furthermore, our research strongly advocates for the collection of denser AVL data, when resources allow, over sparse datasets. The significant reduction in position estimate errors and the improved consistency in metrics like deceleration and speed volatility observed with denser datasets are vital for fine-grained analysis near individual intersections. Overall, this work provides critical insights for researchers and practitioners, guiding the selection of appropriate trajectory reconstruction methods and highlighting the tangible benefits of investing in higher-resolution data for more accurate and realistic analyses.

Future work should study additional approaches, particularly those using machine learning and artificial intelligence. These types of methods were not analyzed in this study since they are generally more computationally expensive and behave less predictably in abnormal scenarios. However, advances in computing power, model stability, and model-based learning methods are making these methods more competitive. Additional work should also examine the predictive power of these algorithms to inform real-time applications when short-term travel times, velocities, or locations need to be predicted between measured points, particularly given the irregular sampling patterns and low frequency inherent in AVL systems. Extending these models to that context would provide additional value beyond the current context of evaluation of historical trends and patterns. 

\newpage
\section*{Acknowledgments}
We gratefully acknowledge the support of Natalia Ruiz Juri, Kenneth Perrine, and the Center for Transportation Research at the University of Texas at Austin for their support in data collection. We would also like to thank the City of Austin and the Capital Metropolitan Transportation Authority, Austin, TX (CapMetro) for their support. 

The authors used Claude Sonnet 4 (Anthropic) for minor editing and table formatting in the preparation of this manuscript.

\section*{Author Contributions}
The authors confirm contribution to the paper as follows: study conception and design: J. Robbennolt; data collection: J. Robbennolt; analysis and interpretation of results: J. Robbennolt, S. Munira; draft manuscript preparation: J. Robbennolt, S. Munira, S.D. Boyles. All authors reviewed the results and approved the final version of the manuscript.

\bibliography{bib/jr_lib, bib/doyun_bib, bib/references}
\bibliographystyle{include/trb}
	
\end{document}